\documentclass[lettersize,journal]{IEEEtran}
\usepackage{amsmath,amsfonts}
\usepackage{algorithmic}
\usepackage{algorithm}
\usepackage{array}
\usepackage[caption=false,font=normalsize,labelfont=sf,textfont=sf]{subfig}
\usepackage{textcomp}
\usepackage{stfloats}
\usepackage{url}
\usepackage[dvipsnames]{xcolor}
\usepackage{hyperref}
\usepackage{verbatim}
\usepackage{graphicx}
\usepackage{cite}
\usepackage{multirow}
\usepackage[flushleft]{threeparttable}
\begin{document}

\title{A Study on the Impact of Face Image Quality on Face Recognition in the Wild}
\author{Na Zhang
\thanks{Na Zhang is with Lane Department of Computer Science and Electrical Engineering at West Virginia University, Morgantown, WV 26506-6109. }
}



\maketitle

\begin{abstract}
Deep learning has received increasing interests in face recognition recently. Large quantities of deep learning methods have been proposed to handle various problems appeared in face recognition. Quite a lot deep methods claimed that they have gained or even surpassed human-level face verification performance in certain databases. As we know, face image quality poses a great challenge to traditional face recognition methods, e.g. model-driven methods with hand-crafted features. However, a little research focus on the impact of face image quality on deep learning methods, and even human performance. Therefore, we raise a question: Is face image quality still one of the challenges for deep learning based face recognition, especially in unconstrained condition. Based on this, we further investigate this problem on human level. In this paper, we partition face images into three different quality sets to evaluate the performance of deep learning methods on cross-quality face images in the wild, and then design a human face verification experiment on these cross-quality data. The result indicates that quality issue still needs to be studied thoroughly in deep learning, human own better capability in building the relations between different face images with large quality gaps, and saying deep learning method surpasses human-level is too optimistic. 
\end{abstract}

\begin{IEEEkeywords}
Face recognition, Face image quality, Deep learning
\end{IEEEkeywords}

\section{Introduction}
\label{intro}
\par We all know that the accuracy of traditional face recognition (FR), e.g. Eigenfaces \cite{turk1991face} and Fisherfaces \cite{belhumeur1997eigenfaces}, is greatly affected by face image quality problems, such as intraclass variations between enrollment and identification stages. Using face images with poor quality can actually degrade face recognition performance. Non-standard lighting or pose and out-of-focus are among the main reasons responsible for the performance degradation. That is why many quality enhancement methods were proposed to try to improve the performance. For example, Hassner \textit{et~al.} \cite{hassner2015effective} used an off-the-shelf detector to detect faces and facial landmarks, and then align the photo with a textured, 3D model of a generic, reference face. Wang \textit{et~al.} \cite{wang2011illumination} performed photometric normalization on face images. One solution, where most researchers commit themselves, is to improve the algorithm itself by making it robust to possible degradation.

\par As the introduction of deep learning (DL) technique, successful development have been obtained on face recognition \cite{schroff2015facenet, lu2015surpassing, wen2016discriminative, liu2017sphereface, wang2018cosface, deng2018arcface}, especially in unconstrained environment, in which the face images contain various face quality challenges, e.g. pose variations, facial expression, varying illumination, large age gap, facial makeup, partial occlusions. Deep learning based face recognition methods can obtain much robust features and outperform the conventional face recognition methods with hand-craft features. Some of these methods claimed that they have achieved human-level performance or even better in face verification on the Labelled Faces in the Wild (LFW) \cite{huang2007labeled} database. The gap between humans and machines seems become narrower. 

\par LFW database is a well-known, widely used, and challenging benchmark for face verification evaluation, which contains 13,233 face images of 5,749 subjects collected from the web. Many deep learning based face recognition methods use this database to evaluate their performance in unconstrained condition. Even though existing face verification accuracy is very close to 100\%, it still remains an argument that claiming surpassing human-level face verification performance is too optimistic. Liao \textit{et~al.} in \cite{liao2014benchmark} figured out that the existing standard LFW protocol is very limited, only 3,000 positive and 3,000 negative face pairs for classification, and fails to fully exploit all the available data. Probably that is why some deep methods can easily reach such high accuracies, even surpass the human-level performance. N. Zhang and W. Deng \cite{zhang2016fine} also proposed several limitations on LFW, like that intraclass variations and interclass similarity sometimes may be ignored by researchers, insufficient matching pairs can not capture the real difficulty of large-scale unconstrained face verification problem. Therefore, it is questionable to say that deep models have touched the limit of LFW benchmark. 

\par For traditional automatic face recognition systems, their performance largely depends on the quality of the face images. Generally speaking, face image quality can be used as a measure metric for their performance. In the early stage, most face images were obtained under controlled environment with proper lighting condition, frontal pose, neutral expression, no or less makeup and standard image resolution, e.g. photos on ID cards. These faces own pretty high quality, thus it is easy for FR systems to achieve extremely high recognition accuracy. However, as the emergency of face data captured under uncontrolled environment (e.g. face images crawled from Internet), these images with low quality significantly degrade recognition accuracy. Some researchers tend to seek for more robust methods, thus deep learning based method was brought in. Different with traditional methods which are model driven, deep learning methods are learning driven which can automatically learn all kinds of faces with different quality problems if enough data are fed into the network. It seems that face image quality become less important for the performance of deep learning based face recognition system. Besides little research specially study the impact of face image quality on deep learning methods.

\par It is well known that face recognition in unconstrained condition is much more difficult due to various changes in face images, e.g. pose variations, illumination changes, varying facial expression, partial occlusion, low resolution, age variations, heavy make-up, etc. Besides, high interclass similarity and large intraclass variation are still two big challenges for face recognition task. Although existing deep models have been trained very well for various quality changes of face images, it is still much more challenging for deep models to recognize faces with quite low quality. Therefore, we raise a question: Does the performance of deep learning based face recognition system still depend on the face image quality? If not, what is the challenge? If so, how it affects? Based on this, we further investigate the impact of face image quality on human performance, and the gap between deep learning and human.

\par In our previous research \cite{guo2018challenge}, we proposed that the face image quality issue is still a grand challenge for deep learning methods. In order to prove this, we developed new face recognition protocols for cross-quality face identification and verification on two public databases, IJB-A \cite{klare2015pushing} and FaceScrub \cite{ng2014data}, and four popular deep models were evaluated under this settings. 
Based on this research, we asked human beings to perform face verification experiment on the faces in unconstrained environment by matching across different face image qualities and further investigate the impact of face image quality on human performance and the distance between human beings and deep learning methods. We also seek to expand previous comparisons \cite{o2007face, kumar2009attribute, phillips2014comparison, o2012comparing, best2014unconstrained, zhou2015naive, phillips2015human, blanton2016comparison} by performing face verification on cross-quality face data in the wild. In our experiment, we focus on face images of extremely difficult levels. These images are chosen from face pairs that the deep model fails to recognize successfully. The evaluation on human performance in face verification discloses that human beings show a different performance with deep learning methods, and saying surpassing human-level is still too optimistic.

\begin{figure*}[!t]
\center
\includegraphics[width=1.0\linewidth]{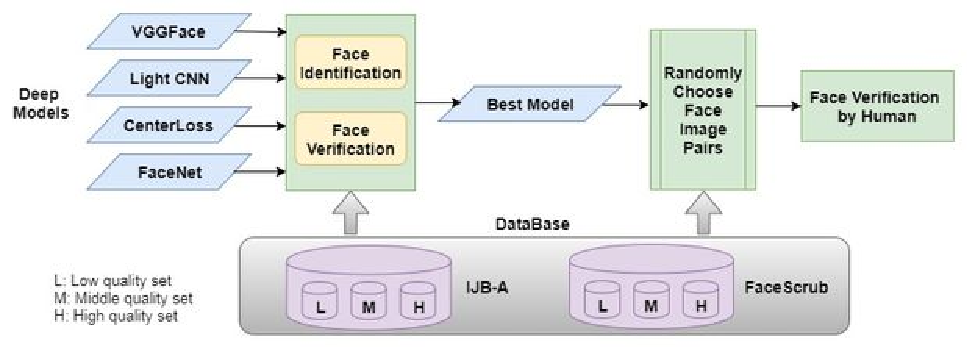}
\caption{Pipeline of our approach.}
\label{fig:pipe}
\end{figure*}

\par The contributions of our work includes:
\begin{itemize}
\renewcommand\labelitemi{\textbullet}
\item as an extension of research \cite{guo2018challenge}, we aim to examine the face recognition performance of deep learning and human beings on cross-quality face images;
\item four pre-trained deep models with high reported accuracy are adopted to perform cross-quality face recognition on two databases, IJB-A and FaceScrub; and the deep model with best recognition performance is chosen to be compared with human beings;
\item human beings perform better than deep learning on face recognition by matching face images with different qualities, especially when the quality gap is large, which also indicates that deep learning method still has a long way to surpass human.
\end{itemize}

\par The paper is organized as follows. In section \ref{related}, we talk about related work on face image quality assessment, human performance in face recognition. In section \ref{approach}, we describe how to choose the best model among four representative deep models. In section \ref{human}, the face verification experiment is performed by human. And section \ref{result} gives an analysis on the results. In section \ref{con}, some interesting discussion and conclusions are drawn.

\section{Related Work}
\label{related}
\subsection{Face Image Quality Assessment}
\par
\begin{figure*}[!t]
\center
\includegraphics[width=0.8\linewidth]{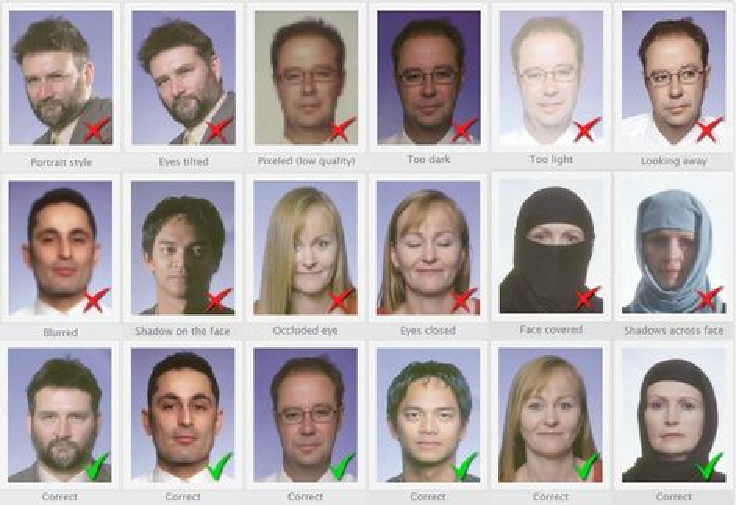}
\caption{Illustration of face images by ISO/IEC 19794-5 standard. Top two rows: incorrect face photos, bottom row: correct face photo.}
\label{fig:idface}
\end{figure*}

 Face image quality is an important factor that apparently affect the performance of traditional face recognition. In practical recognition system, it is usual to choose multiple face images for each subject, hence choosing face images with high quality is a good way to improve recognition accuracy. The approved ISO/IEC standard 19794-5 \cite {bio2007face} specified recommendations for face photo taking for ID card, E-passport and related applications, including instructions for light condition, head pose, facial expression, occlusion, and so on. Figure \ref{fig:idface} shows a few correct and incorrect illustration face images of ISO/IEC 19794-5 standard \cite{isoface}. Face images of bad quality which do not accord with the requirements of the standards is a reason leading to face recognition performance degradation. ISO/IEC 29794-5 \cite{bio2010face} specifies a few methodologies and approaches for computation of quantitative quality scores for facial images by introducing facial symmetry, resolution and size, illumination intensity, brightness, contrast, color, exposure, sharpness, etc.

\par Recently, a few face image quality assessment methods have been proposed. Most existing face image quality assessment methods are based on the analysis of specific facial properties. Yang \textit{et~al.} \cite{yang2004face} introduced a face pose estimation method by a boosting regression algorithm to evaluate face image quality, and applied it in the best shot selection problem to choose the most frontal face from a video sequence. Gao \textit{et~al.} \cite{gao2007standardization} developed a facial symmetry based method for face image quality assessment in which it applies the degree of facial asymmetry to quantify the face quality caused by non-frontal illumination and improper face pose. Nasrollahi and Moeslund \cite{nasrollahi2008face} assesses face quality in video sequence by combining four features (e.g. out-of-plan rotation, sharpness, brightness and resolution) using a local scoring system and weights. Sang \textit{et~al.} \cite{sang2009face} presented several methods for face image quality evaluation. It uses Gabor wavelets as basis features to estimate the facial symmetry and then evaluate the illumination condition and facial pose. Sellahewa \textit{et~al.} \cite{sellahewa2010image} try to measure the face image quality in terms of luminance distortion in comparison to a specified reference face image. Wong \textit{et~al.} \cite{wong2011patch} designed a patch-based face image quality assessment method to choose the 'best' subset of face images from multiple frames of video captured in uncontrolled conditions by quantifying the similarity of a face image to a probabilistic face model, the 'ideal' face. Image characteristics that affect recognition, such as head pose, illumination, shadowing, motion blur and focus change over the sequence, are taken into account. Long and Li \cite{long2011near} designed a quality assessment system to select the best frame from the input video sequence by considering five features including sharpness, brightness, resolution, head pose and expression. The score of each feature is calculated separately, and then the final quality score is obtained by weight fusion of five scores. The image quality assessment model in \cite{chen2011image} assesses the image quality by considering occlusion, face-to-camera distance, pose, expression, uneven illumination measure. 

\par Most of the methods mentioned above apply the artificially defined facial properties and empirically selected reference face images in their assessment process. Some others apply different features, or strategies. Zhang and Wang \cite{zhang2009asymmetry} proposed three asymmetry based face quality measures, which are based on scale insensitive SIFT features. Bharadwaj \textit{et~al.} \cite{bharadwaj2013can} applied Gist and HOG to classify face images into different quality categories that are derived from face matching performance. Raghavendra \textit{et~al.} \cite{raghavendra2014automatic} proposed a scheme for face quality estimation. It first separates frontal faces from non-frontal ones by pose estimation, and evaluate the image quality of frontal faces by analyzing its texture components using Grey Level Co-occurrence Matrix (GLCM), finally quantify the quality using likelihood values obtained using Gaussian Mixture Model (GMM). Chen \textit{et~al.} \cite{chen2015face} proposed a simple and flexible framework in which multiple feature fusion and learning to rank are used.

\subsection{Human Performance in Face Recognition}
\par A lot researchers did pretty much work to evaluate human performance in face recognition. O'Toole \textit{et~al.} \cite{o2007face} did a series of face verification experiments on human and algorithms in which the face images of each pair were taken under different illumination conditions. They found that three algorithms surpassed humans being performance by matching face pairs pre-screened to be "difficult" and six algorithms surpassed humans on "easy" face pairs. Alice J. O'Toole \textit{et~al.} \cite {o2012comparing} compared the performance of humans and machines in face identification task on frontal face images taken under different uncontrolled illumination conditions in both indoor and outdoor settings and with natural variations in a person's day-to-day appearance. In particular, they studied how human beings perform relative to machines as the level of difficultly increases as the variations contributed, such as facial expression, partial occlusion, hair styles and so forth. They concluded that the superiority of machines over humans in the less challenging conditions may indicate that face recognition systems may be ready for applications with comparable difficulty. 

\par Kumar \textit{et~al.} \cite{kumar2009attribute} presented an evaluation of human performance on LFW dataset by following a procedure mentioned in paper \cite{o2007face}. They generated 6,000 image pairs and asked 10 users to label two faces of each pair whether they belong to same person or not. The users were also asked to rate their confidence when labelling. Human performance on LFW is 99.20\%, 97.53\% and 94.27\% when users are shown the original images, tighter cropped images and inverse crops. Human performance is really perfect when the participants are shown the original images. Due to lacking context information, the performance drops when a tighter cropped version of face images are given. It indicates that human can easily use context cues to recognize faces. Besides, the human performance is still wonderful when they are just shown the inverse cropped version (only context information is shown). P. Jonathon Phillips \textit{et~al.} \cite{phillips2014comparison} also did a similar work by matching frontal faces in still and video face images in different difficulty levels (e.g. good, challenging, very challenging). The result showed that algorithms are consistently superior to humans for frontal still faces with good quality, and humans are superior for video and challenging still faces. The result also indicated that humans can use non-face identity cues (e.g. head, body. etc.) to recognize faces. Best-Rowden \textit{et~al.} \cite{best2014unconstrained} analyzed the face recognition accuracies achieved by both machines and humans on unconstrained face data, reported the human accuracy in still images via crowdsourcing on Amazon Mechanical Turk, and first reported human performance on video faces, the YouTube Faces database, which indicated that humans are superior to machines, especially when videos contain contextual cues in addition to the face image.

\par Zhou \textit{et~al.}\cite{zhou2015naive} did a human face verification test in real-world environment on Chinese ID (CHID) benchmark, in which the data were collected offline and specialized on Chinese people. The dataset contains a typical characteristic, age variation including intra-variation (i.e., same person with different ages) and inter-variation (i.e., persons with different ages). The experiment focused on cases their recognition system failed to recognize. The result showed that 90\% cases can be solved by human. Phillips \textit{et~al.} \cite{phillips2015human} expanded the comparison between human and machine from still images and videos taken by digital single lens reflex cameras to digital point and shoot cameras, Point and Shoot Face Recognition Challenge (PaSC). They provided a human benchmark for verifying unfamiliar faces in unconstrained still images at two levels: challenging and extremely-difficult. 100 different-identity image-pair with the highest similarity scores and 100 same-identity image-pair with the lowest similarity scores were selected and 30 users were asked to view two faces of each image-pair side by side and rate on a 1 to 5 scale. The results demonstrated that, in extremely-difficult level, human performance shines relative to algorithms.

\par Austin Blanton \textit{et~al.} \cite{blanton2016comparison} also made a comparison of performance between human and algorithms in face verification on the challenging IJB-A dataset, which includes varying amounts of imagery, immutable attributes,e.g. gender, and circumstantial attributes, e.g. occlusion, illumination, and pose. In their experiment, the participants are asked to show how confident when they decide whether two given faces belong to same subjects or not with six options, which are Certain, Likely, Not Sure, Unlikely, Definitely Not, and Not Visible. The result shows that even for the challenging images in IJB-A, face verification is an easy task for humans.

\par In the past 10 years, pretty a lot researchers studied the performance of humans and machines on face recognition and did all kinds of comparisons between them. In some scenarios, especially "easy" cases, the algorithms perform better, and in other scenarios, like still images in "difficult" levels with various variations and videos, the humans are better. As the fast development of deep learning technique in face recognition, the performance of deep models increase quickly. Quite a lot research reported the surpassing human-level performance on face recognition. Can deep learning technique really gain more excellent performance than human?


\section{Our Approach}
\label{approach}

\begin{figure}[!t]
\center
\includegraphics[width=1.0\linewidth]{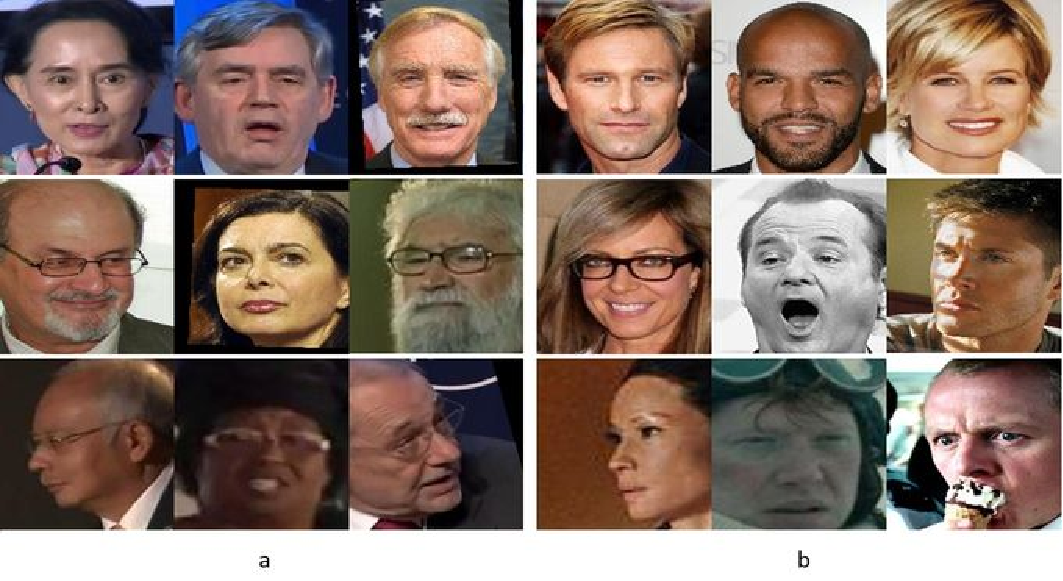}
\caption{Face examples of High (top row), Middle (middle row), and Low (bottom row) quality sets from two databases: (a) IJB-A, (b) FaceScrub.}
\label{fig:example}
\end{figure}

\par Fig. \ref{fig:pipe} shows a whole pipeline of our approach. At the beginning, we partition two popular public databases in the wild, IJB-A \cite{klare2015pushing} and FaceScrub \cite{ng2014data}, into three quality sets (e.g. high quality, middle quality, low quality) separately according the face image quality score. Four famous pre-trained deep models, Light CNN \cite{wu2015light}, FaceNet \cite{schroff2015facenet}, VGGFace \cite{parkhi2015deep}, and CenterLoss \cite{wen2016discriminative}, with high reported accuracy, are chosen to perform face recognition experiments, including face identification and face verification, on cropped faces of the two databases. After that, the deep model with best performance among them is selected by evaluating their performance. And the face images that the best model fails to recognize successfully are filtered as the data to be used in our well-designed human verification experiments. Human beings are asked to perform face verification experiment by matching across different face image qualities and then the result is evaluated to further examine whether face image quality changes can impact the performance of human beings, how, and what is the gap between deep model and human. In the experiment, we focus on extremely difficult level of face images, i.e., matching low to high quality sets. These images are chosen from face pairs that deep model fails to recognize successfully.

\begin{table}[!t]
\center
\caption{Face images distribution on IJB-A and FaceScrub.}
\label{tab:dis}
\begin{tabular}{ c |l r r}
\hline\noalign{\smallskip}
& Quality Set & \# Images & \# Subjects  \\
\noalign{\smallskip}\hline\noalign{\smallskip}
\multirow{3}{*}{IJB-A} & High & 1,543 & 500 \\
&Middle & 13,491 & 483 \\
&Low & 6,196 & 489 \\
\hline
\multirow{3}{*}{FaceScrub} & High & 10,089 & 530 \\
&Middle & 10,444 & 530 \\
&Low & 362 & 232 \\
\noalign{\smallskip}\hline
\end{tabular}
\end{table}

\begin{figure*}[!t]
\center
\includegraphics[width=1.0\linewidth]{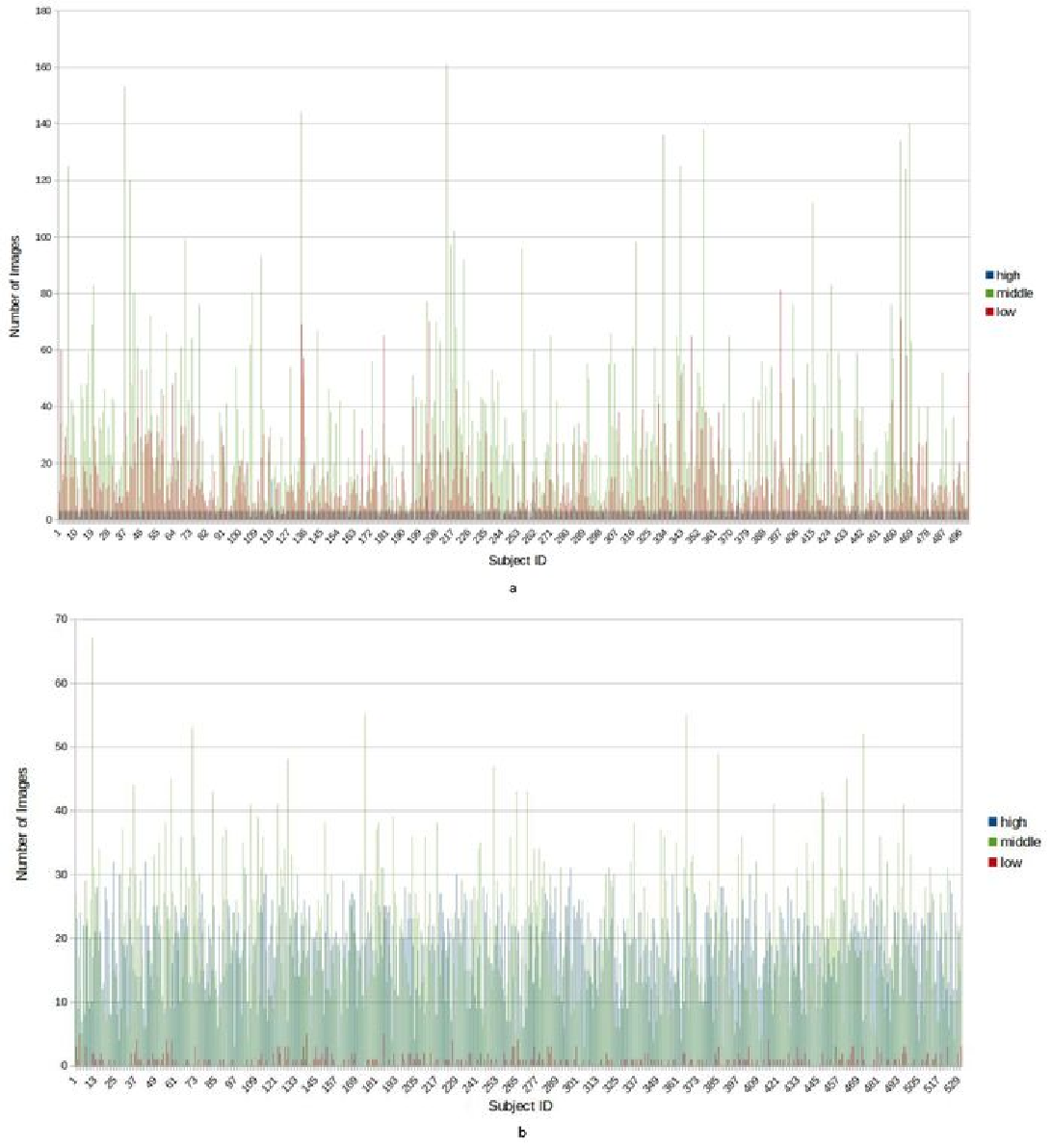}
\caption{Distribution of High, Middle, and Low quality sets for each subject on (a) IJB-A and (b) FaceScrub databases. \textbf{Best view in color}}
\label{fig:dbdis}
\end{figure*}

\subsection{Face Image Quality}
\par Although LFW is very popular for face recognition in the wild, there still exists some limitations, like the standard LFW protocol contains limited number of pairs, which causes insufficient exploration on various quality issues, e.g. pose variations, lighting condition, low resolution. Therefore, face image quality changes maybe the key issue in unconstrained face recognition. In order to have a better understanding of the face image quality, we are first to examine the distribution of different face qualities in the data and the impact of the distribution on face recognition performance.

\par The face image quality is evaluated by considering specific facial properties, like resolution, pose angle, illumination parameters, or occlusion. We adopt a method proposed in \cite{chen2015face} to measure and quantify the quality of every face image. This method tries to compare the relative qualities of each face pairs and then use the relative relationship to train a ranking based model to learn the quality score. The generated quality score, which is between 0 and 100, is used as the indicator of face image quality. The higher the quality score is, the better quality the face image has. According to the score of face image, the database is divided into three subsets, i.e., high quality, middle quality and low quality sets. In our study, high quality set is selected as the gallery set, and middle, low quality sets as probe set separately, and then to perform face recognition on four deep models.

\subsection{Database Preparation}
\par We evaluate the performance of face recognition with matching across different face image quality sets on two public face databases, IJB-A \cite{klare2015pushing} and FaceScrub \cite{ng2014data}. IJB-A, the IARPA Janus Benchmark A (IJB-A) database, is a publicly available media in the wild dataset containing a total of 21,230 face images of 500 subjects with manually localized face images. It is more challenging for face recognition. This dataset contains full pose variation, joint use for face recognition and face detection benchmark, wider geographic variation of subjects, protocols supporting both open-set identification (1:N search) and verification (1:1 comparison), an optional protocol that allows modelling of gallery subjects and ground truth eye and nose locations. FaceScrub was created by building face dataset that detects faces in images returned from searching for public figures on the Internet, followed by automatically discarding those not belonging to each queried person. It comprises a total of 106,863 face images of 530 celebrities with about 200 images per person. It contains 55,306 face images of 265 males and 51,557 face images of 265 females. 

\par All face images in both databases are estimated by the face image quality assessment method \cite{chen2015face} and quality scores are calculated for each face image. According to these scores, we divide the two databases into three different quality sets. Table \ref {tab:dis} shows the distribution of three quality sets on the two databases. The quality score is between 0 and 100. Image quality scores in high quality set are greater than or equal to 60. Scores in middle quality set are greater than or equal to 30 and less than 60. And scores in low quality set are less than 30. Fig. \ref{fig:example} gives some face examples of high, middle, and low quality sets from the two databases. Images with high quality are those frontal faces with high resolution, proper light condition, no occlusion. Images with low quality are those with big pose, dark light condition, or partial occlusion. And images with middle quality are those cases between the two situations.

\par For IJB-A database, we find that quite a lot subjects in high quality set have less than three images. To ensure the gallery, i.e., high quality set, has enough target faces (at least three), we choose a few images from middle quality sets with higher scores to the high quality set. From Fig.\ref{fig:dbdis} (a), it is easy to notice that most subjects in high quality set have three images. The middle quality set contains the most images (63.55\%), and low quality set also contains pretty much (29.19\%). However, FaceScrub database owns many images with pretty good quality, about 70\% images with high quality and 25\% with middle quality. In order to match the size of IJB-A, a shortened version of FaceScrub is generated by randomly selecting images from each subject in high and middle quality sets. Finally, the subset of FaceScrub contains a total of 20,895 images of 530 subjects as shown in table \ref {tab:dis}. From Fig.\ref{fig:dbdis} (b), we can see the shortened FaceScrub still has quantities of face images with pretty good quality.

\subsection{Deep Models}
\par Light CNN \cite{wu2015light}, FaceNet \cite{schroff2015facenet}, VGGFace \cite{parkhi2015deep}, and CenterLoss \cite{wen2016discriminative} are four popular deep models that have reported very high accuracies (LightCNN: 99.33\%, FaceNet: 99.63\%, VGGFace: 98.95\%, and CenterLoss: 99.28\%) on LFW for face verification. Light CNN \cite{wu2015light} is a light framework to learn a 256-D face representation on the large-scale face data with massive noisy labels. It is efficient in computational costs and storage spaces. FaceNet \cite{schroff2015facenet} can directly learn a mapping from input face images to a compact 128-D Euclidean space in which the Euclidean distance indicates face similarity. VGGFace \cite{parkhi2015deep} is inspired by \cite{simonyan2014very}. It is a 'very deep' network with a long sequence of convolutional layers. CenterLoss \cite{wen2016discriminative} uses two loss functions, softmax and center loss, to train the deep model. The center loss can learn a center of deep features for each class to reduce the intra-class variations and enlarge the inter-class differences.

\subsection{Choose Model with Best Performance}
\par To avoid any bias in training stage, we use the pre-trained deep models to perform cross-quality face identification and verification experiments on three types (high, middle, and low) of quality sets from IJB-A and FaceScrub databases. By evaluating the performance, the model with best performance is selected.

\subsubsection{Face Identification}
\begin{figure}[!t]
\center
\includegraphics[width=1.0\linewidth]{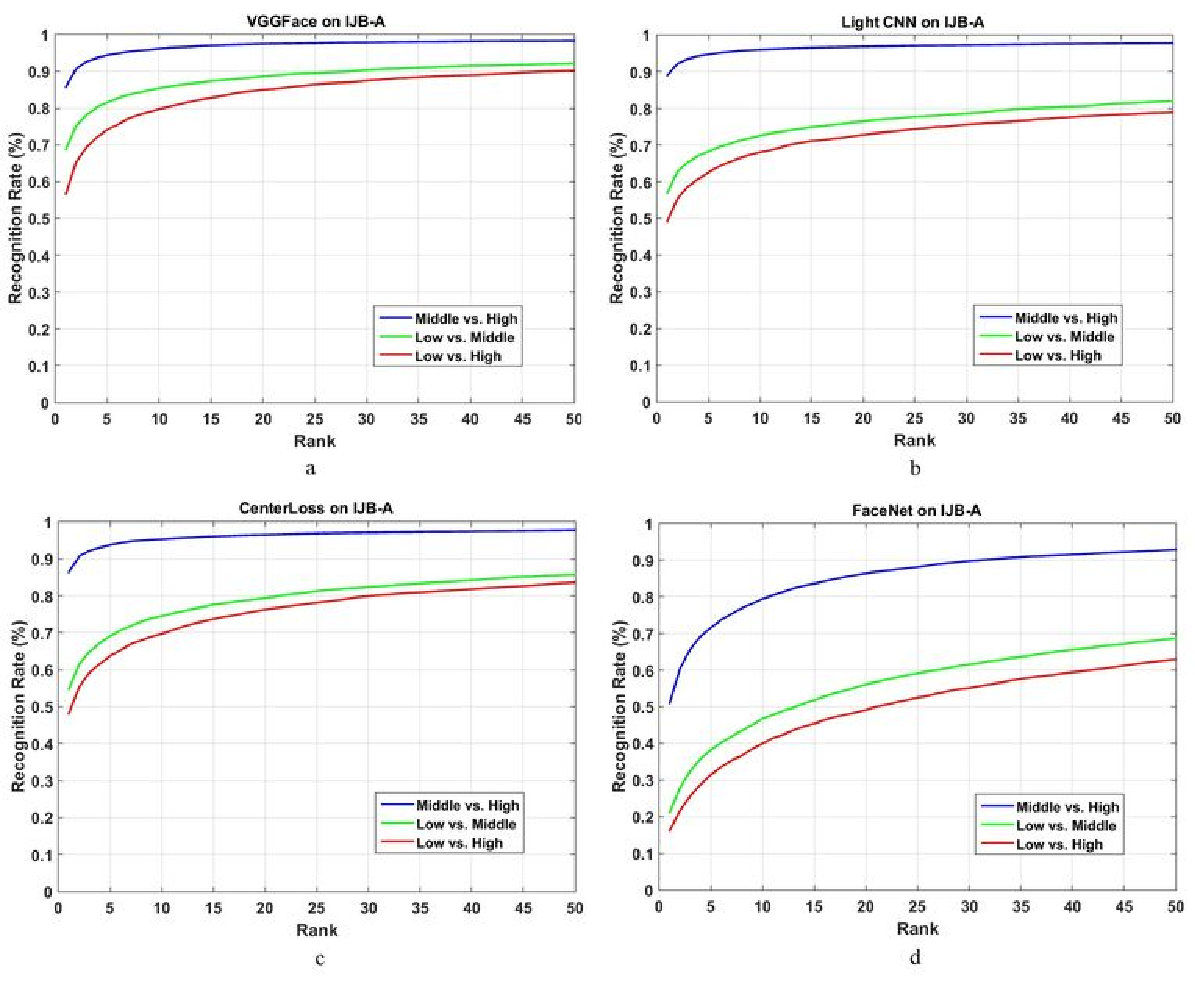}
\caption{CMC of face identification experiments by matching different quality images using (a) VGGFace, (b) Light CNN, (c) CenterLoss, and (d) FaceNet on IJB-A. \textbf{Best view in color}}
\label{fig:fiden1}
\end{figure}

\begin{figure}[!t]
\center
\includegraphics[width=1.0\linewidth]{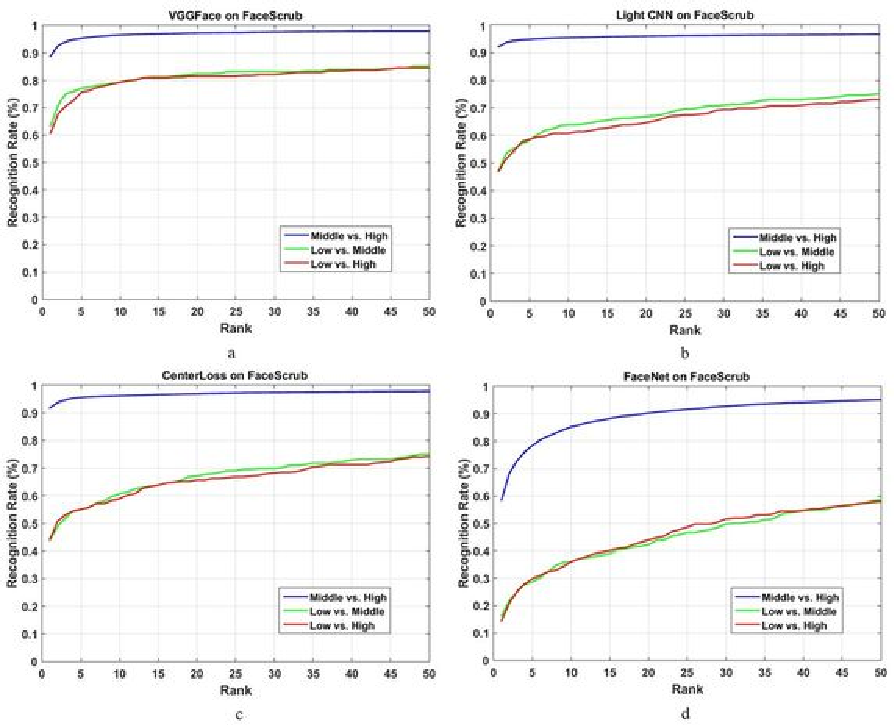}
\caption{CMC of face identification experiments by matching different quality images using (a) VGGFace, (b) Light CNN, (c) CenterLoss, and (d) FaceNet on FaceScrub. \textbf{Best view in color}}
\label{fig:fiden2}
\end{figure}

\par Face identification aims to recognize the person from a set of gallery face images and find the most similar one to the probe sample. For each database, we design three groups of experiments, and in each group the matching faces is across different quality sets. The first one is low to high matching in which low quality set is designed as query images and high quality set is gallery images. The second one is middle to high matching in which middle quality set is query images and high quality set is gallery images. And the third one is low to middle matching in which query images come from low quality set and gallery images are from middle quality set. Deep features of three quality sets from four deep models on IJB-A and FaceScrub are extracted and Cosine Similarity Score is adopted to calculate the similarity score of each face pair. The performance of four models is measured by Cumulative Match Curve (CMC) \cite{bolle2005relation} on two databases as shown in Fig.\ref{fig:fiden1} and Fig.\ref{fig:fiden2}. It is easily to find that the performance of matching from middle to high quality set is much better than the other two matches for all deep models. The performance of matching from low to middle is slightly better than that of matching from low to high for most cases. The reason probably is that the difference between low and high quality faces is larger than the difference between low and middle quality faces. In general, VGGFace has the better result than the other three models, and FaceNet performs the worst.

\subsubsection {Face Verification}
\par Face verification aims to determine whether a given pair of face images or videos belongs to the same person or not. Considering that the performance of low to high and low to middle quality sets are nearly similar, only low to high and low to middle cases are performed in face verification experiment. Low and middle quality sets of each database are set as query images separately and high quality set as gallery images. Finally, about 18,978 positive pairs and 9,541,450 negative pairs in the case of matching low to high quality sets, and 41,642 positive pairs and 20,774,971 negative pairs in the case of matching middle to high quality sets on IJB-A database are generated, and also 6,676 positive pairs and 3,645,542 negative pairs in the case of matching low to high quality sets, and 193,745 positive pairs and 105,175,771 negative pairs in the case of matching middle to high quality sets on FaceScrub database are generated.

\begin{table*}[!t]
\center
\caption{Face verification rersult on four deep models.}
\label{tab:far}
\begin{tabular}{ l|l|ccc|ccc}
\hline\noalign{\smallskip}
\multirow{2}{*}{\textbf{Database}} & \multirow{2}{*}{\textbf{Deep Model}} 
& \multicolumn{3}{c}{\textbf{Low vs. High}} & \multicolumn{3}{c}{\textbf{Middle vs. High}} \\
& & FAR=0.01 & 0.001 & 0.0001 & 0.01 & 0.001 & 0.0001 \\
\noalign{\smallskip}\hline\noalign{\smallskip}
\multirow{5}{*}{IJB-A} & VGGFace & \textbf{0.605} & 0.367 & 0.194 & 0.858 & 0.675 & 0.491 \\ 
& LightCNN & 0.566 & \textbf{0.402} & \textbf{0.269} & \textbf{0.905} & \textbf{0.808} & \textbf{0.678} \\
& CenterLoss & 0.521 & 0.313 & 0.164 & 0.859 & 0.692 & 0.499 \\
& FaceNet & 0.257 & 0.100 & 0.033 & 0.586 & 0.330 & 0.165 \\ 
& Gabor & 0.037 & 0.006 & 0.001 & 0.200 & 0.112 & 0.064 \\
\hline
\multirow{5}{*}{Shortened FaceScrub} & VGGFace & \textbf{0.595} & \textbf{0.389} & \textbf{0.231} & 0.837 & 0.662 & 0.468 \\
& Light CNN & 0.503 & 0.330 & 0.148 & 0.896 & 0.811 & \textbf{0.668} \\
& CenterLoss & 0.493 & 0.341 & 0.215 & \textbf{0.914} & \textbf{0.814} & 0.652 \\
& FaceNet & 0.219 & 0.075 & 0.019 & 0.633 & 0.350 & 0.162 \\
& Gabor & 0.022 & 0.003 & 0.001 & 0.082 & 0.027 & 0.010 \\
\hline
\end{tabular}
\end{table*}

\begin{figure}[!t]
\center
\includegraphics[width=1.0\linewidth]{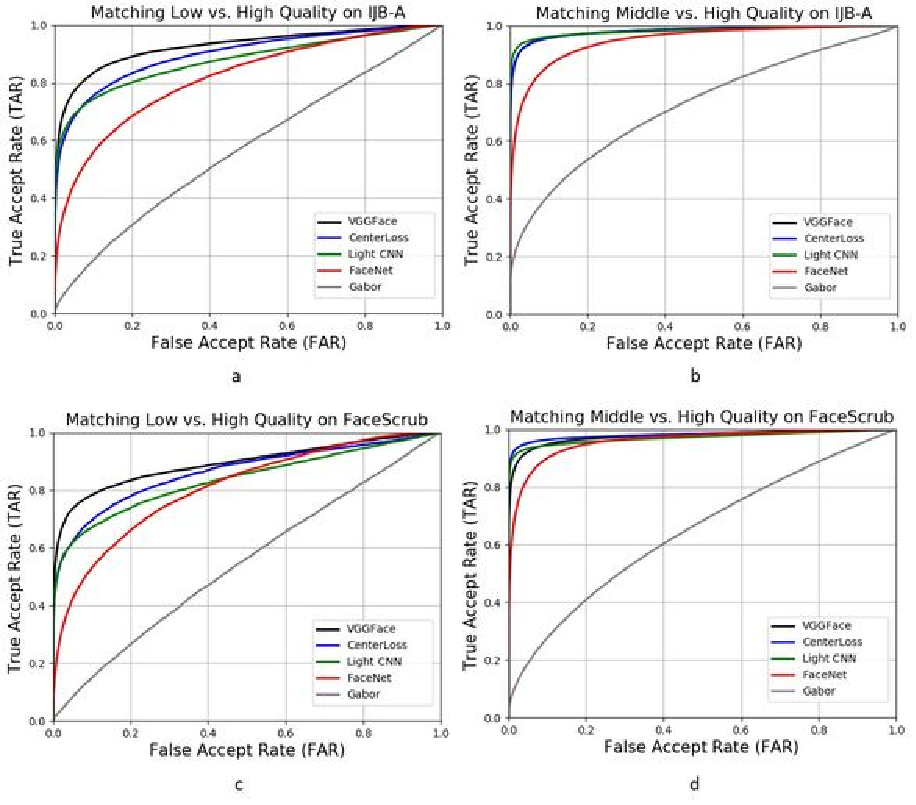}
\caption{ROC of face verification experiment by matching (a) Low vs. High Quality on IJB-A, (b) Middle vs. High on IJB-A, (c) Low vs. High on FaceScrub, and (d) Middle vs. High on FaceScrub. \textbf{Best view in color}}
\label{fig:roc}
\end{figure}

\par In the face verification experiment, we construct a similarity matrix in which the row presents one query image, the column indicates one gallery image and the value in the matrix shows cosine similarity score between two face images of the corresponding row and column. Simultaneously, a similarity mask matrix is built in which the row still indicates one query image and the column indicates one gallery image. The difference between the two matrices is the values. In similarity mask matrix, the values have only two types. -1 means that two face images in the corresponding row and column is a positive pair and 127 means negative pair. We still adopt Cosine Similarity Score to show how similar two faces are and then calculate verification accuracies with respect to FAR=0.01, 0.001 and 0.0001 (FAR: false accept rate) as presented in table \ref{tab:far}, and also give Receiver Operating Characteristic curves (ROC) in Fig. \ref{fig:roc}. The result of verification using Gabor feature is set as a baseline to be compared. We can see that the performance of Gabor feature is the worst. There is a big gap between Gabor features and deep features. For matching middle to high quality sets experiment, Light CNN and CenterLoss has the best performance on IJB-A and FaceScrub separately. And in low to high experiment, VGGFace performs best on FaceScrub, and better than others in FAR=0.01 case on IJB-A.

\par By analyzing the results of face identification and verification experiments, we can see that, on IJB-A, VGGFace has the best performance in low to high experiment, Light CNN is the best in middle to high experiment, and on FaceScrub, VGGFace gains the highest accuracy in low to high experiment, CenterLoss performs best in middle to high experiment.

\section{Face Verification Experiment by Human}
\label{human}
\par In this face verification experiment, we use the best model chosen from previous face identification and verification experiments, and try to find the decision boundary for these positive and negative face pairs based on the best model. Then we randomly select a certain number of face pairs that the best model fails to recognize and perform human verification experiment on the selected face pairs.

\par Since our goal is to examine how well the human performance on face verification comparing to algorithms, we mainly focus on face verification task in extremely difficult level, matching low quality set to high quality set. From previous experiments, it is easy to find that VGGFace has the greatest performance on IJB-A and FaceScrub databases in matching low to high experiment. Hence we choose a number of face image pairs of low to high quality set on IJB-A and FaceScrub databases based on VGGFace model to do human face verification experiment.

\subsection{Get Decision Boundary}
\par
\begin{figure}[!t]
\center
\includegraphics[width=1.0\linewidth]{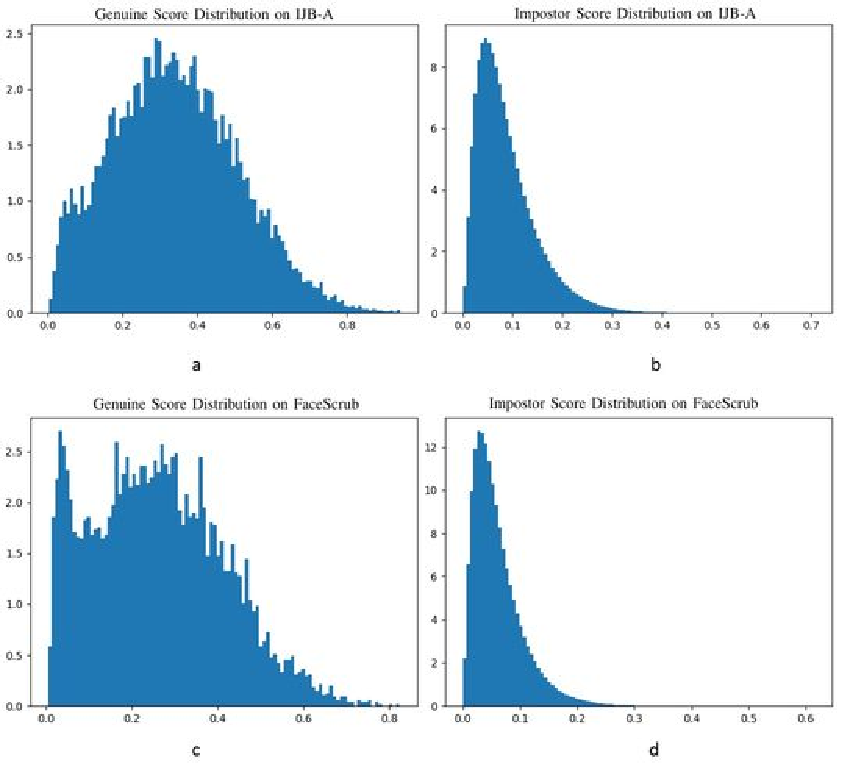}
\caption{Genuine and impostor score distribution on IJB-A and FaceScrub. (a) genuine score distribution on IJB-A, (b) impostor score distribution on IJB-A, (c) genuine score distribution on FaceScrub, and (d) impostor score distribution on FaceScrub.}
\label{fig:hs}
\end{figure}

\begin{figure}[!t]
\center
\includegraphics[width=1.0\linewidth]{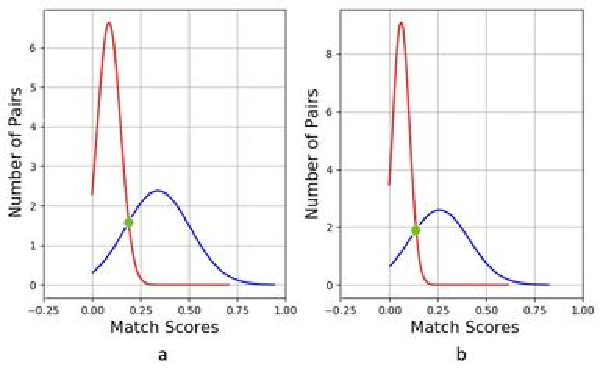}
\caption{Genuine (blue line) and impostor (red line) matching score distribution on (a) IJB-A and (b) FaceScrub. The threshold value is from the match score of green dot shows. \textbf{Best view in color}}
\label{fig:fit}
\end{figure}

 We generate the statistical distributions of genuine and impostor matching scores of all positive and negative pairs on the two databases to find the decision boundaries. Fig. \ref{fig:hs} shows the statistical distributions of genuine and impostor scores on both databases. And then the distributions are fitted as Gaussian distribution illustrated in Fig. \ref{fig:fit}. Finally, the thresholds, 0.188 for IJB-A and 0.138 for FaceScrub, are easily obtained. 

\subsection{Choose Genuine and Impostor Pairs}
\par Based on the thresholds, genuine and impostor pairs can be easily selected. Those face images that VGGFace fails to recognize successfully are chosen, so the genuine pairs whose matching scores are less than the threshold value and the impostor pairs whose matching scores are greater than or equal to the threshold are filtered from two databases. Since context information in face image can give people some useful cues to recognize the identities \cite{kumar2009attribute}, the original images are not directly used in the experiment. We adopt a cropped version of original face images from VGGFace. Besides, those pairs that the face images are wrongly or improperly aligned or cropped are manually removed to ensure that those pairs in the human experiment do not contain some technical errors caused by the factors that , not image quality. And then we randomly select 100 positive pairs and 100 negative pairs from the cleaned pairs, put them together and randomly permute them. Finally, a total of 400 pairs for two databases are obtained. In this case, the verification rate of deep model VGGFace is 0\% correct.

\subsection{Participants and Tool}
\par
\begin{figure}[!t]
\center
\includegraphics[width=1.0\linewidth]{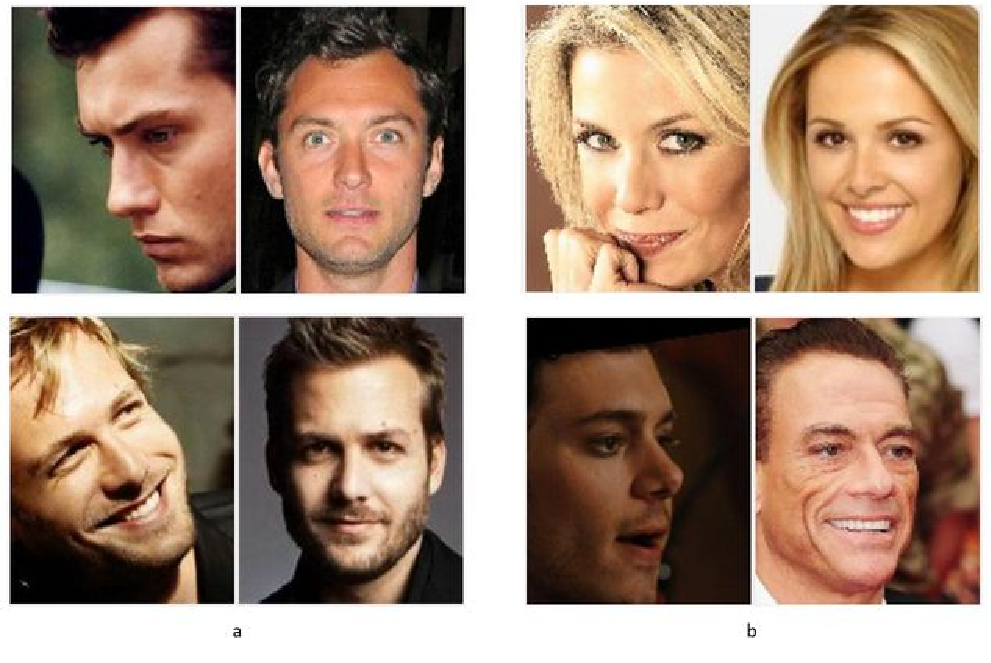}
\caption{Samples of face images pairs: (a) two genuine pairs, and (b) two impostor pairs. }
\label{fig:tool}
\end{figure}

 We design a face verification experiment performed by humans. In the experiment, a total of 20 participants, 14 males and 6 females, are asked to view 400 face image pairs and give their choice on whether the two faces in each given pair belong to same person or not. A part of them (as indicated in table \ref{tab:group}) have much experience on face image quality analysis, some ones just know about it and others have no background. For convenience, a tool is designed to assist participants during experiment. Fig. \ref{fig:tool} shows some samples of face pairs shown in the tool. Left is two positive pairs and right are two negative ones. 

\begin{table}[!t]
\center
\caption{Details on three groups of participants.}
\label{tab:group}
\begin{tabular}{ ccccl}
\hline\noalign{\smallskip}
Groups & \# Male & \# Female & In Total & Description \\
\noalign{\smallskip}\hline\noalign{\smallskip}

All & 14 & 6 & 20 & \begin{tabular}{@{}l@{}} \end{tabular}\\
Group1 & 2 & 1 & 3 & \begin{tabular}{@{}l@{}} Have much experience on \\ face  quality \end{tabular} \\
Group2 & 2 & 2 & 4 & \begin{tabular}{@{}l@{}}Worked on some facial  \\ image analysis tasks\end{tabular} \\
Group3 & 10 & 3 & 13 &\begin{tabular}{@{}l@{}} Have no background  \end{tabular} \\
\noalign{\smallskip}\hline
\end{tabular}
\end{table}

\subsection{Experiment Procedure}
\par 100 positive pairs and 100 negative pairs are randomly selected for each database. These 200 pairs are divided into four subsets randomly with same size, i.e., 50 pairs. A total of eight subsets are generated in the end. All participants are asked to check the pairs one by one for each subset on the designed tool and make the decision. After finishing one subset, participants are advised to check next subset after a pretty good rest which makes them work on this task with full of energy. All participants have unrestricted time to finish this experiment.

\section{Experiment Results and Analysis}
\label{result}
\par
\begin{figure}[!t]
\center
\includegraphics[width=1.0\linewidth]{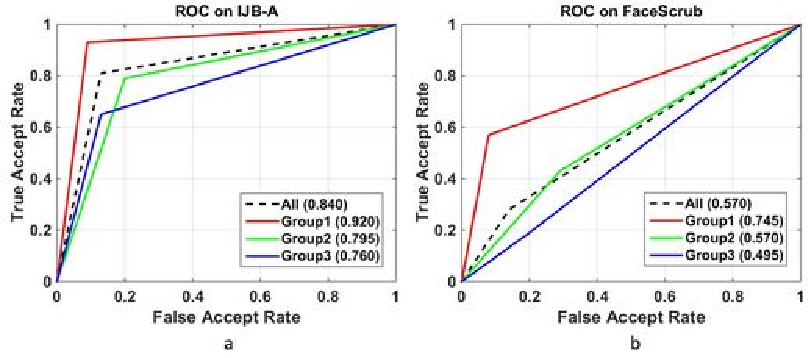}
\caption{ROC of human face verification experiment (a) IJB-A and (b) FaceScrub.}
\label{fig:human}
\end{figure}

 All participants are grouped into three sets as indicated in Table \ref{tab:group} according to their background on image quality analysis. 3 persons (2 males and 1 female) have quite a lot experience on face quality understanding and analysis. 4 individuals (2 males and 2 females) have ever worked on related topics, and the remaining (10 males and 3 females) have little background. We also analyzed all participants as one group. Most of them are students. Majority voting technique is adopted to deal with the final results of these four groups. If the number in the group is even, one subject in it will randomly removed and just odd number of subjects are considered. Table \ref{tab:cmt} and \ref{tab:cmt1} gives the confusion matrix results including positive and negative accuracies in both actual and predicted cases on IJB-A and FaceScrub databases. ROC curves are also drawn in Fig. \ref{fig:human}.

\begin{table}[!t]
\center
\caption{Confusion Matrix Result on IJB-A Database.}
\label{tab:cmt}
\begin{tabular}{ccccc}
\hline\noalign{\smallskip}
\multicolumn{2}{c}{\multirow{2}{*}{\textbf{IJB-A:All}}} & 
\multicolumn{2}{c}{Predicted} & \multirow{2}{*}{Accuracy} \\
\multicolumn{2}{c}{} & Positive & Negative & \\
\noalign{\smallskip}\noalign{\smallskip}
\multirow{2}{*}{Actual} & Positive & $81\%$ & $19\%$ & \multirow{2}{*}{$84\%$} \\
& Negative & $13\%$ & $87\%$ &\\
\hline
\multicolumn{2}{c}{\multirow{2}{*}{\textbf{IJB-A:Group1}}} & 
\multicolumn{2}{c}{Predicted} & \multirow{2}{*}{Accuracy} \\
\multicolumn{2}{c}{} & Positive & Negative & \\
\multirow{2}{*}{Actual} & Positive & $93\%$ & $7\%$ & \multirow{2}{*}{$92\%$} \\
& Negative & $9\%$ & $91\%$ & \\
\hline 
\multicolumn{2}{c}{\multirow{2}{*}{\textbf{IJB-A:Group2}}} & 
\multicolumn{2}{c}{Predicted} & \multirow{2}{*}{Accuracy} \\
\multicolumn{2}{c}{} & Positive & Negative & \\
\multirow{2}{*}{Actual} & Positive & $79\%$ & $21\%$ & \multirow{2}{*}{$79.5\%$} \\
& Negative & $20\%$ & $80\%$ & \\
\hline 
\multicolumn{2}{c}{\multirow{2}{*}{\textbf{IJB-A:Group3}}} & 
\multicolumn{2}{c}{Predicted} & \multirow{2}{*}{Accuracy} \\

\multicolumn{2}{c}{} & Positive & Negative & \\
\multirow{2}{*}{Actual} & Positive & $65\%$ & $35\%$ & \multirow{2}{*}{$76\%$} \\
& Negative & $13\%$ & $87\%$ & \\
\hline
\end{tabular}
\end{table}

\begin{table}[!t]
\center
\caption{Confusion Matrix Result on FaceScrub Database.}
\label{tab:cmt1}
\begin{tabular}{ccccc}
\hline\noalign{\smallskip}
\multicolumn{2}{c}{\multirow{2}{*}{\textbf{FaceScrub: All}}} & 
\multicolumn{2}{c}{Predicted} & \multirow{2}{*}{Accuracy} \\
\multicolumn{2}{c}{} & Positive & Negative & \\
\noalign{\smallskip}\noalign{\smallskip}
\multirow{2}{*}{Actual} & Positive & $28\%$ & $72\%$ & \multirow{2}{*}{$57\%$} \\
& Negative & $14\%$ & $86\%$ & \\
\hline 
\multicolumn{2}{c}{\multirow{2}{*}{\textbf{FaceScrub:Group1}}} & 
\multicolumn{2}{c}{Predicted} & \multirow{2}{*}{Accuracy} \\
\multicolumn{2}{c}{} & Positive & Negative & \\
\multirow{2}{*}{Actual} & Positive & $57\%$ & $43\%$ & \multirow{2}{*}{$74.5\%$} \\
& Negative & $8\%$ & $92\%$ & \\
\hline 
\multicolumn{2}{c}{\multirow{2}{*}{\textbf{FaceScrub:Group2}}} & 
\multicolumn{2}{c}{Predicted} & \multirow{2}{*}{Accuracy} \\
\multicolumn{2}{c}{} & Positive & Negative & \\
\multirow{2}{*}{Actual} & Positive & $43\%$ & $57\%$ & \multirow{2}{*}{$57\%$} \\
& Negative & $29\%$ & $71\%$ & \\
\hline 
\multicolumn{2}{c}{\multirow{2}{*}{\textbf{FaceScrub:Group3}}} & 
\multicolumn{2}{c}{Predicted} & \multirow{2}{*}{Accuracy} \\
\multicolumn{2}{c}{} & Positive & Negative & \\
\multirow{2}{*}{Actual} & Positive & $19\%$ & $81\%$ & \multirow{2}{*}{$49.5\%$} \\
& Negative & $20\%$ & $80\%$ & \\
\hline
\end{tabular}
\end{table}

\par By analyzing the results, we can easily find that the performance of human on IJB-A and FaceScrub is more excellent than VGGFace (best among the four deep models), although very high accuracy on LFW benchmark is achieved. There still exists a clear gap between human performance and machine recognition especially in the real-world setting. Real-world face recognition has much more diverse criteria, like big pose angle, poor illumination condition, and large facial occlusion, than we treated in previous recognition benchmarks. And data quality plays an important role in the performance of algorithms. Wider and more arbitrary range of changes like pose, illumination, expression, occlusion, resolution, age variation, heavy make-up of face images are most common factors which influence the system's performance. However, it still lacks a sufficient investigation on these cross factors, and also lacks an efficient method to handle them clearly and comprehensively. Large amount of face data with these factors are needed to assist us to build better models to improve recognition performance. 

\par We also find that people who have much experience in face recognition perform better than those who have not. What is interesting is that people have higher accuracy in recognition of negative pairs than that of positive pairs. The reason may be that it is hard for people to recognize that the two faces belong to same subject for positive pairs since the quality of face in query set is much low, but for negative pairs, it is much easier to view two faces as negative (different persons). Besides, we find that the accuracies on FaceScrub are lower than IJB-A. The reason may be that the quality of faces in query set (low quality set) on FaceScrub is much lower than that on IJB-A. The quality scores of face images can also prove this.

\section{Discussion and Conclusion}
\label{con}
\par It is obvious that face image quality plays an important role in model-driven face recognition systems. Faces with bad quality can directly degrade the accuracy of face recognition. The main reason may be that most face recognition methods in the early stage try to build the models that are used to extract hand-craft features, and nearly all data are collected in controlled conditions with standard lighting, fixed head pose, proper facial expression, etc. These data fails to contain various or mixed qualities of face images. And the built models are sensitive to face quality changes. In order to improve the accuracy, some research focus on designing face image quality enhancement methods, like deblurring \cite{pan2014deblurring}, pose correction \cite{hassner2015effective}, and photometric normalization \cite{wang2011illumination}. Another solution is to develop more robust algorithm to possible degradation. The brought of deep learning technique into face recognition field gives an clear direction to further development.

\par In our previous research \cite{guo2018challenge}, we explored the impact of face image quality on deep learning based face recognition in unconstrained environment. Practically, the performance of deep neural networks can be largely improved by feeding various of face data with different qualities in training stage. Since the deep networks have almost learnt all kinds of face images with different qualities, they may keep in mind certain connections between them on some level. Hence deep learning based face recognition system can obtained more robust features than traditional face recognition methods. However, in fact, face image quality still has an influence on the accuracy of face recognition, although the deep networks have seen large quantities of face images. For example, in face identification evaluation on four deep models, it is easy for deep models to identify the correct subject in matching faces from middle to high qualities, but difficult in matching from low to high, which shows that deep models can recognize faces whose quality changes are big to some degrees, but not too huge. Therefore, more robust deep learning methods than existing ones are still needed to be able to recognize faces with large quality gaps.

\par The influence of face image quality on human performance were further explored. We designed a face verification experiment by human beings on cross-quality face data, IJB-A and FaceScrub, by matching from low to high qualities, which is the hardest one.
The human performance on IJB-A and FaceScrub are more excellent than the best model, VGGFace. Human outperform deep learning methods largely. The result indicts that there still exists a clear gap between human and machine performance in face recognition in unconstrained environment. Human beings own the capability in recognizing face images with large quality gaps. Besides, all participants were grouped into three categories according to their background on face image quality analysis, and the performance of each group were analyzed too.

\bibliographystyle{IEEEtran}
\bibliography{Face_Image_Quality_on_Face_Recognition}

\end{document}